\documentclass[10pt,letterpaper]{article}
\usepackage[top=0.85in,left=2.75in,footskip=0.75in]{geometry}

\usepackage{amsmath,amssymb,booktabs}

\usepackage{changepage}

\usepackage{textcomp,marvosym}

\usepackage{cite}

\usepackage{nameref,hyperref}

\usepackage[right]{lineno}

\usepackage[nopatch=eqnum]{microtype}
\DisableLigatures[f]{encoding = *, family = * }

\usepackage[table]{xcolor}

\usepackage{array}

\newcolumntype{+}{!{\vrule width 2pt}}

\newlength\savedwidth

\raggedright
\usepackage[aboveskip=1pt,labelfont=bf,labelsep=period,justification=raggedright,singlelinecheck=off]{caption}

\makeatletter
\renewcommand{\@biblabel}[1]{\quad#1.}
\makeatother

\usepackage{lastpage,fancyhdr,graphicx}
\usepackage{epstopdf}
\fancyheadoffset[L]{2.25in}
\fancyfootoffset[L]{2.25in}
\usepackage{underscore}
\usepackage{arydshln}

\begin{document}
\vspace*{0.2in}

\begin{flushleft}
{\Large
 \textbf\newline{Beyond surface form: A pipeline for semantic analysis in Alzheimer's Disease detection from spontaneous speech} 
}
\newline
\\
Dylan Phelps\textsuperscript{1,2,*},
Rodrigo Wilkens\textsuperscript{3},
Edward Gow-Smith\textsuperscript{2},
Lilian Hubner\textsuperscript{4},
Bárbara Malcorra\textsuperscript{4},
César Rennó-Costa\textsuperscript{5},
Marco Idiart\textsuperscript{6},
Maria-Cruz Villa-Uriol\textsuperscript{1,2},
Aline Villavicencio\textsuperscript{2,3,5}
\\
\bigskip
\textbf{1} Healthy Lifespan Institute, The University of Sheffield, Sheffield, United Kingdom
\\
\textbf{2} School of Computer Science, The University of Sheffield, Sheffield, United Kingdom
\\
\textbf{3} Institute of Data Science and Artificial Intelligence, The University of Exeter, Exeter, United Kingdom
\\
\textbf{4} Faculty of Linguistics, Pontifícia Universidade Católica do Rio Grande do Sul, Porto Alegre, Brazil
\\
\textbf{5} Bioinformatics Multidisciplinary Environment, Digital Metropolis Institute, Federal University of Rio Grande do Norte, Natal, Brazil
\\
\textbf{6} Institute of Physics, Federal University of Rio Grande do Sul, Porto Alegre, Brazil
\bigskip

* drsphelps1@sheffield.ac.uk

\end{flushleft}
\section*{Abstract}

Alzheimer’s Disease (AD) is a progressive neurodegenerative condition that adversely affects cognitive abilities, including language comprehension and production. These language-related changes can be automatically identified through the analysis of outputs from linguistic assessment tasks, such as picture description. Language models show promise as a basis for screening tools for AD, but their limited interpretability poses a challenge in distinguishing true linguistic markers of cognitive decline from surface-level textual patterns. To address this issue, we examine how surface form variation affects classification performance, with the goal of assessing the ability of language models to represent underlying semantic indicators. We introduce a novel approach where texts surface forms are transformed by altering syntax and vocabulary while preserving semantic content. The transformations significantly modify the structure and lexical content, as indicated by low BLEU and chrF scores, yet retain the underlying semantics, as reflected in high semantic similarity scores. This allows us to isolate the effect of semantic information on AD detection, finding models perform similarly to if they were using the original text, with small $\pm$0.1 deviations in macro-F1. We also investigate whether language from picture descriptions retains enough detail to reconstruct the original image using generative models. We found that visual elements are poorly preserved, and image-based transformations add substantial noise, reducing text alignment and classification accuracy. Our methodology provides a novel way of looking at what features influence model predictions, and allows the removal of possible spurious correlations. We find that just using semantic information, language model based classifiers can still detect AD. This work shows that difficult to detect semantic impairment can be identified, addressing an overlooked feature of linguistic deterioration, and opening pathways for better early detection systems.

\section*{Introduction}

Globally, both the proportion and the number of people 65 years or older are growing rapidly, with 16\% of the population, or 1.6 billion people, expected to be over 65 by 2050 \cite{worldpopulation2022}. Age is the highest risk factor for Alzheimer's Disease (AD), with 1 in 14 over 65 being affected by the condition \cite{wittenberg_projections_2020}. With this aging population and the increasing number of people living with AD, there is a need for methods of diagnosis that are more efficient and that capture the changes early.

AD is a clinical condition characterized by progressive cognitive decline. Subtle changes in speech and language often provide early indicators of impairment \cite{forbes-mckay_detecting_2005} and can provide information on the nature of deterioration \cite{ahmed_connected_2013}. The effects of AD on memory interact with language abilities, with both the access to words (the lexicon) and their meaning (semantics) being affected by the disease \cite{rodrigues_por_2024, coffey_its_2024}. Additionally, some effect on syntax has been observed \cite{saffran_quantitative_1989, thompson_patterns_1997}, however, it has been argued that this effect is due to underlying semantic impairment \cite{reilly2011language}.

The current battery of tests used during diagnosis includes extensive speech and language assessments, which have proven valuable for identifying the deterioration of linguistic and other cognitive constructs \cite{hernandez2018computer, sanborn2022automated}. One tool generally included in batteries of linguistic diagnosis tests is the `picture description task'. Within this task, a participant should describe or tell a story depicted in a fairly simple image or set of images, and their responses can be recorded and further analysed for symptoms of AD \cite{forbes-mckay_detecting_2005, hubnerbale, luz_alzheimers_2020}. In this context the semi-automatic processing of the speech or transcription has been investigated for diagnosis with feature based methods extracting lexical and syntactic information from the surface form \cite{petti_systematic_2020}, and many recent works building upon language models such as BERT \cite{devlin_bert_2019, balagopalan_bert_2020} and GPT-2 \cite{radford_language_2019, li_gpt-d_2022}, and further using Large Language Models (LLMs), such as GPT4 \cite{b_t_performance_2024}. 

However, while feature based methods explicitly incorporate lexical and syntactic features such as pronoun-noun ratio \cite{bittner_changes_2022}, lexical frequency \cite{almor_noun-phrase_1999}, and repetition \cite{8612831} \cite{petti_systematic_2020}, language model–based methods are more difficult to interpret due to the opaque nature of the distributed representations within language models \cite{rogers_primer_2020}. As they can use various features of the text, it remains uncertain whether they depend primarily on superficial characteristics or can truly grasp the more nuanced linguistic structures beneath, and how robust they are to surface form changes.

As semantic degradation has been shown to be a key factor in AD-related language impairment \cite{reilly2011language}, it is important to explore the role of semantic information in automatic AD detection, and separate it from the surface level indicators. With this understanding, future work can improve the performance of both semantics only classifiers and classifiers using all features of a text.




\subsection*{Research questions}


We propose a pipeline to disentangle semantic features from surface form features in AD patients' speech by transforming the original text. Specifically, we use generative language models to standardize the surface characteristics of the texts, therefore, producing text that still has a high level of semantic similarity with the original, allowing for classification models to be trained that focus only on the semantic information. In this study, these transformations consist of restructuring the content into a storyboard format that sequentially organizes key information into text-only scenes and producing summaries of different lengths.

Additionally, as an extreme case study where all surface elements of the text are removed, we propose using models that can create images from text descriptions (text-to-image) and, vice versa (image-to-text) to extract only the semantic component from the original texts.

We evaluate our methodologies by comparing the original data to our created data using two similarity metrics for the surface form (BLEU \cite{papineni_bleu_2002}, chrF \cite{popovic_chrf_2015}) and a semantic similarity metric (embedding cosine similarity \cite{reimers_sentence-bert_2019}). We also use the transformed data to train a BERT classifier \cite{devlin_bert_2019} for automatic AD detection, to compare the performance on a downstream task.

In this work, we aim to address three research questions:
\begin{enumerate}
    \item Can we use LLMs to standardise the AD transcripts surface form, removing lexico-syntactic indicators, whilst retaining the semantic information?
    \item How do language model based AD classifiers perform when focusing solely on the semantic information?
    \item What does this tell us about the importance of semantic information in automatic AD detection?
\end{enumerate}




By answering these questions, we hope to explore what information language models use in the detection of AD and propose methods for extracting semantic information so that the semantic impairment of patients with AD can be explored further.

Overall, our results show that our text-to-text methods can alter the surface form sufficiently, leading to texts with low surface form similarity (BLEU $<0.1$, chrF $<0.5$) and high semantic similarity ($>0.6$). We also show that this transformed text can still be used to automatically classify between AD patients and the control group with small $\pm$0.1 macro-F1 differences between the best performing transformations and the original data. Conversely, we show that current text-to-image and image-to-text generation algorithms introduce to much noise, which hinders the accurate preservation of semantic information. This results in low semantic similarity and greater performance variability.

The small changes in AD detection performance for our text-to-text pipeline show that language models are robust to changes in the surface form and therefore are capable of making classifications based on the underlying semantic information. This adds to previous results that show how comprehension is the main language function impacted by AD, and allows for further work that improves our understanding of semantic impairment in AD, and language models' ability to detect it. This increased understanding will allow the development of automatic AD detection systems that more accurately use semantic features complementary to the commonly used surface-form features, increasing overall performance.

\section*{Background}

\subsection*{Effect of AD on language}

Dementia affects several types of memory, especially episodic and semantic ones. Language can also be affected \cite{weiner_language_2008, ahmed_connected_2013}, as besides being a complex symbolic system, it interacts with other cognitive constructs, like memory types, executive functions, and attention. For instance, studies on connected speech have shown that low speech connectedness in AD is associated with poorer semantic memory performance, which, in turn, impacts episodic memory capacity \cite{malcorra_low_2021}. Psycholinguistic studies have shown that certain psycholinguistic features are sensitive to reduced semantic ability in Alzheimer’s, namely word concreteness, age of acquisition, animacy, and frequency, to name some of these features \cite{raling_judging_2017}.  

Semantic memory loss is a marker of early Alzheimer's disease-related neurodegeneration in older adults 
\cite{vonk_semantic_2020}.  Moreover, there is a correlation between clinical dementia measures such as the Mini-Mental State Exam (MMSE) \cite{folstein_mini-mental_1975, tombaugh_mini-mental_1992} with various linguistic aspects\cite{fraser_linguistic_2016, kave_severity_2018, malcorra_exploring_2024}. For instance, the study developed by Kavé and Dassa (2018) showed a correlation between MMSE scores with lexical features and information units in a narrative production based on pictures in their AD group but not in the control group. 

In addition, AD might also impact syntax. Some studies \cite{saffran_quantitative_1989, thompson_patterns_1997} have identified patients who produce agrammatic speech, whilst, conversely, \cite{kempler_syntactic_1987} showed that syntactic ability can be preserved.
It is widely discussed whether the effects on syntax seen in AD patients are due to memory and semantic impairment \cite{reilly2011language}. It is not an easy task to try to split apart what relates to syntax versus semantics when it comes to sentence and discourse processing \cite{ehrlich_ideational_1997, sajjadi_abnormalities_2012}. Furthermore, cognitive deficits, like those in semantic memory, may interfere with both syntactic and semantic processing at different levels and also depending on the disease stage \cite{nasiri_investigating_2022}.


\subsection*{Language model-based AD classification}

Automatic language assessment methods have been refining their ability to assess language for AD, and much of this work has focused on extracting linguistic features, such as lexical frequency and noun-pronoun ratio, and training classifiers based on these. More recently, however, we have seen the rise of transformer \cite{vaswani_attention_2017} language models such as BERT \cite{devlin_bert_2019} and GPT \cite{radford_language_2019, brown_language_2020}, in addition to their high level of performance in general tasks, these models have also demonstrated promising results when applied to tasks in the medical domain
\cite{lee_biobert_2020}. 

These models have subsequently been applied to automatic AD detection tasks, with many studies using the Alzheimer's Dementia Recognition through Spontaneous Speech Challenge (ADReSS) shared task dataset \cite{luz_alzheimers_2020}. ADReSS presents a balanced dataset consisting of audio recordings and transcripts of spontaneous speech in response to the Cookie Theft picture description task \cite{goodglass1983assessment}. Work on this task using fine-tuned language models, specifically BERT, has been shown to produce good performance \cite{balagopalan_bert_2020, agbavor_predicting_2022}, generally better than classifiers based on using the audio data \cite{cummins_comparison_2020}, and those utilizing linguistic features \cite{balagopalan_comparing_2021}.

With the advent of LLMs such as GPT-4 \cite{openai_gpt-4_2024}, some initial work has applied these models, directly out-of-the-box, to AD detection from spontaneous speech tasks, with limited success \cite{b_t_performance_2024, yang_gpt-4_2023}. Furthermore, some research, which is closest to ours in this paper, has attempted to extract semantic features from the ADReSS transcripts by prompting GPT-4 to score each transcript on a number of linguistically relevant metrics, such as discourse impairment and semantic paraphasias \cite{heitz_linguistic_2025}. Overall, they find that the using GPT-4 derived metrics alongside traditional metrics for classification improves performance.

\section*{Materials and methods}

\subsection*{Datasets}

We use two datasets of responses to picture description tasks. We use one dataset in English and one in Portuguese, allowing us to compare the performance of the proposed method in two different languages. A summary of the size of the datasets and the average length of each entry is show in Table \ref{tab:datasets}.

\begin{table}[!ht]
\centering
\caption{
{\bf Dataset statistics}}
\begin{tabular}{lccccc}
\toprule
Dataset & \multicolumn{3}{c}{Number of Texts} & \multicolumn{2}{l}{Transcript Length} \\ \cmidrule{2-6} 
 & AD & Control & Total &  AD & Control \\ \midrule
Dog Story & 23 & 116 & 139 &  105($\pm$58) & 127($\pm$54) \\
ADReSS & 78 & 78 & 156 &  90($\pm$51) & 105($\pm$50) \\ \bottomrule
\end{tabular}
\begin{flushleft} Count of the number of texts for each group in each dataset and the mean($\pm$std) transcript length, by number of words, in the AD and control groups.\end{flushleft}
\label{tab:datasets}
\end{table}

\subsubsection*{Dog Story dataset}
``The Dog Story'', a subtest of the Battery for Language Assessment in Ageing \cite{hubnerbale}, is a dataset of transcribed responses, in Brazilian Portuguese, where participants were asked to tell a story based on seven scenes. The dataset has transcripts from 139 participants. Of those participants, 23 had been diagnosed with AD, whilst 116 were from the control group. As a result, the dataset is very imbalanced, which presents challenges when using it to train and evaluate a classification model. The measures taken to handle this imbalance are described in the next section.

\subsubsection*{Alzheimer's Dementia Recognition through Spontaneous Speech}

In addition to evaluating the Dog Story task, we also evaluate the Alzheimer's Dementia Recognition through Spontaneous Speech (ADReSS) dataset \cite{luz_alzheimers_2020}. ADReSS is a set of transcripts of patients' and controls' spontaneous speech in response to the Cookie Theft picture task \cite{goodglass1983assessment}. The dataset is balanced for diagnosis, age, and gender, with a total of 156 participants, 78 diagnosed with AD and 78 controls.  This allows results to be compared with the larger body of work on AD recognition through language, due to the widespread use of ADReSS.

\subsubsection*{Ethics Statement}
Both datasets used in the study, Dog Story and ADReSS, are obtained via third parties. Data was fully deidentified and included speech recordings, transcripts and basic demographic data. The Dog Story dataset was collected between 2014 and 2023 to analyze speech in the elderly for individuals with cognitive decline. Ethics approval was granted by the Pontificia Universidade Catolica do Rio Grande do Sul (Ref: 53696221.4.1001.5336). 
DementiaBank reviewed and approved our data access request to use the ADReSS dataset, which is a subset of the Pitt Corpus \cite{becker_natural_1994}, collected between 1983 and 1988 by the University of Pittsburgh Alzheimer's Research Program. 

For both datasets, all enrolled participants provided informed written consent. The University of Sheffield provided ethical approval and confirmed that this work meets the conditions consented to by participants in both datasets (Ref: 067658).

\subsection*{Transformation pipeline}

\begin{figure}[!h]
\includegraphics[width=\textwidth]{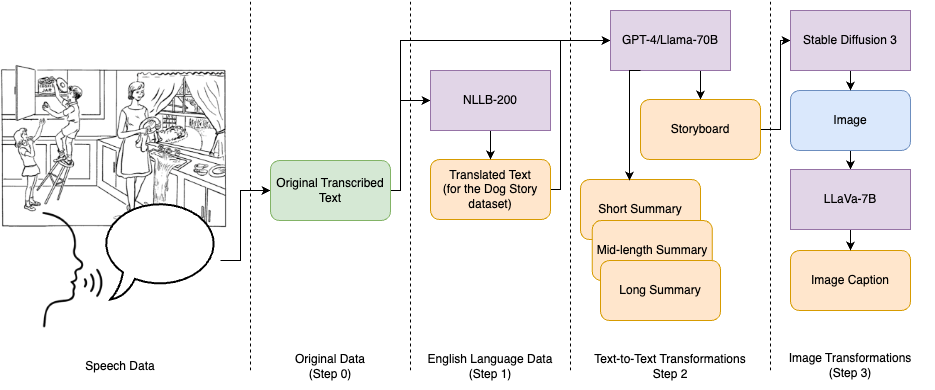}
\caption{{\bf Transformation Pipeline}
Our pipeline to transform the text. We present 3 main steps: translation, text-to-text transformation, and image and caption generation. The models used in the pipeline are shown in purple, the pipeline outputs used in the experiment are shown in orange.}
\label{semextpip}
\end{figure}

We use LLMs to transform the transcribed speech of participants responding to the picture description task, to alter the surface form, whilst maintaining the semantic information. To do this, we propose a pipeline of methodologies, with each stage producing text that abstracts the meaning from the original descriptions. The pipeline can be seen in Fig \ref{semextpip}, and an example of the output from the pipeline applied to a random example from the Dog Story dataset can be found in Appendix \ref{app:example-output}. The code used to implement the pipeline can be found on GitHub \cite{github-repo}.

Step 1 in the pipeline translates non-English data into English using a generative language model. This applies only to the Dog Story dataset (originally in Portuguese). We use NLLB-200 \cite{team_no_2022} for this translation, enabling all datasets to utilize the same state-of-the-art English language models in subsequent steps.

Steps 2a and 2b both perform text-to-text transformations using generative language models. For the Dog Story dataset, we employ GPT-4o \cite{openai_gpt-4o_2024}, the current SOTA model. Due to data agreement restrictions for ADReSS, we cannot use 3rd party models, therefore, we use a 6-bit quantized version of Llama-3.3 70B \cite{grattafiori_llama_2024}, which is optimized for our hardware (1 80GB H100 GPU). All the prompts used across steps 2 and 3 can be seen in Table \ref{tab:prompts}.

\begin{table}[h!]
\caption{
{\bf Generation prompts}}
\begin{tabular}{c p{0.21\linewidth} p{0.64\linewidth}}
\toprule
Step & Transformation & Prompt \\
\midrule
2 & Short Summary & Summarise the following text into a one sentence summary. Just output the summary and no other information. \\ 
2 & Medium Summary & Summarise the following text into a concise summary. Just output the summary and no other information. \\ 
2 & Long Summary & Summarise the following text into a long summary containing as much information as possible. Just output the summary and no other information. \\ 
2 & Storyboard & Transform any text you are given into key story scenes. Focus on the most important story moments. Break down complex actions into separate scenes if needed. Just output the storyboard and no other information. \\ 
3 & Image Caption & Describe in detail what is happening in the image. \\ 
\bottomrule
\end{tabular}
\begin{flushleft}
Prompts used with  the (multimodal) Large Language Models for generation at each step in our pipeline
\end{flushleft} 
\label{tab:prompts}
\end{table}

In step 2a, we transform the text into structured ``Storyboards'', while in step 2b, we generate summaries at three different lengths. For step 2b, we produce 3 different summary lengths to investigate how the similarity and downstream performance change as the model is forced to remove more information to create shorter texts. Both methodologies will keep the key semantic information described in the original text while altering the surface form. Additionally, the transformation in step 2a should improve the text's formatting for image generation in step 3.

The final transformation (step 3 in Fig \ref{semextpip}) converts the storyboard descriptions into images using a text-to-image model, then regenerates text using an image-to-text model. This process leverages the fact that images represent semantic content without syntactic information, allowing us to extract meaning independently from the original surface forms. After finding similar performance between third-party and local models for image and caption generation, we use StableDiffusion-3XL \cite{esser_scaling_2024} for image generation and LLaVa-8B \cite{liu_visual_2023} for caption generation across both datasets.

\section*{Metrics}
\label{sec:experiments}

To assess to what extent the proposed method successfully changes surface forms while maintaining the relevant semantic information, we adopt similarity measures targeting both surface forms and semantic information. Additionally, we evaluate the extrinsic impact of these transformations in a classification task of AD vs. Control groups.

\subsection*{Similarity scoring}

We use a number of similarity score metrics to compare the outputs from the different types of transformation directly to each other and the original texts. These metrics verify to what extent the surface form of the generated texts is substantially different from the originals while maintaining the underlying information linked to AD relevant for classification. 

For surface form similarity, we use the similarity metrics BLEU\cite{papineni_bleu_2002} and chrF\cite{popovic_chrf_2015}. BLEU uses the proportion of matching n-grams (sequences of $n$ adjacent words) to calculate a measure of lexical overlap between pairs of texts, whereas chrF performs a similar calculation but instead at the character n-gram level (sequence of n characters). We use both metrics, as whilst BLEU can capture word choice and phrase structure, it can also miss partial matches such as from morphological inflection or from different word forms (e.g. ``quick'' vs ``quickly'' scores 0 BLEU, but $>0$ chrF). Both metrics produce scores between 0 and 1, with higher scores implying a closer match. chrF will typically be higher for any pair of sentences as character-level matching is less strict, with scores $<0.4$ indicating a poor match, whilst a BLEU score of $<0.2$ indicates similarly poor quality. Therefore, if both metrics are low, we can conclude that there is very little overlap between the surface forms resulting from the different transformations.

For semantic similarity, we use the cosine similarity between SentenceBERT \cite{reimers_sentence-bert_2019} embeddings. SentenceBERT has been shown to effectively encode semantic information into dense vector representations, positioning semantically similar texts closely within the embedding space. Texts with comparable meanings have vectors separated by small angles, resulting in high cosine similarity values. Cosine values range between 0 and 1, with $>0.6$ generally indicating a high level of similarity and thus semantic information being conserved across the transformation, while values $<0.4$ indicate low similarity and low semantic conservation. In our experiments, we use all-mpnet-base-v2, which is a variant of MPNet \cite{song_mpnet_2020} fine-tuned for high-quality sentence representations, to create the embeddings for each of our texts.

\subsection*{Classification}

After verifying that the transformations substantially altered the surface form of the original texts, we now aim to determine whether we can automatically classify AD from the information after each of the transformations. To this end, our classification is based on a standard methodology using BERT, which has been shown to have a high performance on the ADReSS dataset by Balagopalan et al. \cite{balagopalan_bert_2020}.

For each step of the pipeline for the Dog Story dataset, we perform 5-fold cross-validation and train our BERT model on the remaining 80\% of the dataset for each fold. The ADReSS dataset contains ready-defined train and test splits that we have adhered to in our evaluation. When training and evaluating the data in English, we use the bert-base-uncased model \cite{devlin_bert_2019}, and for Portuguese, we use BERTimbau Base \cite{souza_bertimbau_2020}. 

Each classification model is trained for 10 epochs with early stopping to prevent overfitting. To calculate the overall score for each step, we average the accuracy (on each class individually) and macro-F1 achieved on each of the folds. Due to the large class imbalance in the Dog Story dataset, we use a weighted loss function when training the classifier on this dataset, using the `\texttt{calculate_class_weight}' function from scikit-learn\cite{scikit-learn}. Additionally, we use macro-F1 as the main metric for our analysis. 

For each transformation, we perform 10 paired classification runs using identical data splits and random seeds. Statistical significance is assessed using a paired Wilcoxon signed-rank test.

\section*{Results}

We discuss the results obtained both in terms of the intrinsic similarities between the original texts and their transformations and also of the extrinsic performance of AD classification. The goal is to determine if the relevant signals for the latter can still be accessed even with transformations, and we use the former to measure substantial changes.

\subsection*{Similarity results}


Results with all three similarity measures are compatible with our expected results for the transformations on both datasets. We calculate the mean similarity between corresponding pairs of sentences in the original and transformed texts. Table \ref{table:sim-scores} presents the mean similarity scores between each transformation and the original text (or the translated English text for the Dog Story dataset). Pairwise similarity metrics between each transformation are shown and discussed in Appendix \ref{sec:conf-matrices}. 

\begin{table}[!ht]
\centering
\caption{
{\bf Similarity scores}}
\begin{tabular}{llrrr:rrr}
\toprule
Step & Transformation & \multicolumn{3}{c:}{Dog Story} & \multicolumn{3}{c}{ADReSS} \\ \cmidrule(lr){3-8} 
 &  & \multicolumn{1}{c}{chrF} & \multicolumn{1}{c}{BLEU} & \multicolumn{1}{l:}{Cosine} & \multicolumn{1}{c}{chrF} & \multicolumn{1}{c}{BLEU} & \multicolumn{1}{l}{Cosine} \\ \midrule
2 & Short Summary & 0.39 & 0.03 & 0.69 & 0.40 & 0.03 & 0.60  \\
2 & Medium Summary & 0.47 & 0.06 & 0.74 & 0.43 & 0.03 & 0.59  \\
2 & Long Summary & 0.35 & 0.05 & 0.76 & 0.17 & 0.02 & 0.67  \\
2 & Storyboard & 0.26 & 0.05 & 0.68 & 0.27 & 0.03 & 0.66  \\
3 & Image Description & 0.25 & 0.01 & 0.39 & 0.23 & 0.01 & 0.45  \\
\bottomrule
\end{tabular}
\begin{flushleft} The similarity scores compared to original data for the surface form (BLEU and chrf) and semantic (Cosine) metrics. The results shown are average values across the examples in each dataset.
\end{flushleft}
\label{table:sim-scores}
\end{table}

\subsubsection*{The Dog Story dataset}

Firstly, for the Dog Story dataset, all of the transformations in the pipeline produce texts with low surface form similarity, as can be seen by the low BLEU scores. The score for each transformation when paired with the translated text is $<0.06$, which shows an extremely low level of lexical overlap between the two texts.

For semantic similarity, the mean cosine similarity scores are relatively high ($\geq0.68$) between the translated text and all of the step 2 text-to-text transformations (Table \ref{table:sim-scores} Cosine). In contrast to the BLEU scores, this shows that much of the semantic information from the original text has been carried through into the transformed data. However, when looking at the similarity compared to the step 3 image captions, a pronounced drop to a cosine similarity score of $0.39$ can be seen, suggesting that in transforming between modalities (from text to image to text) some of the relevant information may be lost. This may be due to the text lacking sufficient detail to reconstruct the salient elements of the original image, or possibly because current multi-modal models aren't sophisticated enough to preserve the relevant information.

\subsubsection*{ADReSS dataset}

The similarity scores on the ADReSS dataset follow a similar trend, with low surface form (BLEU $\leq0.03$; chrF $\leq0.43$ and high semantic similarity ($\geq0.59$) between the transformed data and the original text. However, in general, we see slightly lower semantic similarity scores, which may be a side effect of using the (smaller) local models on this dataset compared with the GPT-4o for the Dog Story data. Again, we see lower similarity with the image caption, $0.45$, suggesting that the loss of information through the image generation is not unique to any one dataset.

\subsection*{Classification results}

\subsubsection*{Dog Story dataset}

Table \ref{tab:dog-results} shows both the accuracy per class and the macro-F1 scores achieved when a BERT classifier was trained at each step of the transformed Dog Story dataset. All metrics are the mean obtained across 10 runs, each performed with a different seed; each run adopts a 5-fold cross-validation, with the metrics averaged across the folds. P-values show are calculated using a paired Wilcoxon signed-rank test.


\begin{table}[!ht]
\caption{\bf Results of classification on Dog Story dataset}
\centering
\begin{tabular}{clccc}
\toprule
Step & Transformation & macro-F1 & \multicolumn{2}{c}{Accuracy per class} \\
\cmidrule(lr){4-5}
 & & & Positive (AD) & Negative (C) \\
\midrule
0 & Original Portuguese & 0.616 & 0.283 & 0.923 \\ \hdashline
1 & Translated English & 0.640$^{*}$ & 0.330 & 0.922 \\
2 & Short Summaries & 0.499 & 0.091 & 0.922 \\
2 & Medium Summaries & 0.547 & 0.170 & 0.917 \\
2 & Long Summaries & 0.647$^{**}$ & 0.300 & 0.949 \\
2 & Storyboard & \textbf{0.662}$^{**}$ & \textbf{0.413} & 0.900 \\
3 & Image Description & 0.523 & 0.091 & \textbf{0.968} \\
\bottomrule
\end{tabular}
\vspace{0.5em}
\\
\begin{flushleft}Mean metrics across 10 runs. Best results for each metric shown in bold. macro-F1 scores that are a statistically significant improvement over the original baseline are marked with $^*$ for p$<$0.05, and $^{**}$ for p$<$0.01. \end{flushleft}
\label{tab:dog-results}
\end{table}

We achieve an F1 score of 0.616 when training the BERTimbau Base model on the original Portuguese data. This performance increases to 0.640 when translating the data to English and using BERT (bert-base-uncased) model for classification.

Our summarized texts show a range of performance, with the short and medium length summaries achieving lower performance than the original data in Portuguese and English, with F1 scores of 0.499 and 0.547, respectively. Our other step 2 transformations, namely the long summaries and storyboards, achieve higher performance than the translated English data, achieving 0.647 and 0.662 F1. The imbalance in the data is visible in the $\geq0.9$ negative accuracy for all the classifiers and the relatively lower positive class accuracy.

Finally, we see a substantial decrease in performance when training on the step 3 data, with the model only achieving a 0.523 F1 score. Looking further, we can see that the mean positive class accuracy is less than 0.1, which indicates that the model nearly always predicts the majority negative class. 

\subsubsection*{ADReSS dataset}

Table \ref{tab:adress-results} shows the mean metrics across 10 runs when a classifier is trained on the data from each of the transformations.


\begin{table}[!ht]
\caption{\bf Results of classification on ADReSS dataset}
\centering
\begin{tabular}{clccc}
\toprule
Step & Transformation & macro-F1 & \multicolumn{2}{c}{Accuracy per class} \\
\cmidrule(lr){4-5}
 & & & Positive (AD) & Negative (C) \\
\midrule
0 & Original English & \textbf{0.802} & \textbf{0.821} & 0.783 \\ \hdashline
2 & Short Summaries & 0.695 & 0.696 & 0.700 \\
2 & Medium Summaries & 0.668 & 0.558 & 0.788 \\
2 & Long Summaries & 0.702 & 0.600 & \textbf{0.817} \\
2 & Storyboard & 0.694 & 0.717 & 0.675 \\
3 & Image Description & 0.527 & 0.592 & 0.475 \\
\bottomrule
\end{tabular}
\vspace{0.5em}
\\
\begin{flushleft}Mean metrics across 10 runs. Best results for each metric shown in bold. macro-F1 for the original text is significantly better than all transformations (Wilcoxon p$<$0.01). \end{flushleft}\label{tab:adress-results}
\end{table}

On the ADReSS dataset, similar trends to the Dog Story dataset are obtained with the transformations as the Long Summaries and Storyboards achieve similar macro-F1 scores of 0.702 and 0.717, higher than any of the other transformations. Moreover, the Storyboards maintain the same balance of performance for the two classes as the Original data, all the other non-image transformations result in more accuracy for the Negative class. Finally, although none of the transformations in isolation can reach the 0.802 macro-F1 score achieved when using the original data to train the classifier, they are still much higher than the random accuracy of $0.5$.

Additionally, we see the same pattern: the image description transformation is far behind the others, with a macro-F1 score of 0.527, close to the random baseline score.

\section*{Discussion}


The results obtained confirm that while our transformations reduce the surface form (or lexico-syntactic) similarities, they seem to maintain the more latent (or semantic) similarities. While the former are consistently low compared to the original text, with a BLEU $< 0.06$ for all transformations for both datasets, the latter remain high, and the transformations that score highest on classification, Long Summaries and Storyboards also have the highest semantic similarity. By calculating the Pearson Correlation Coefficient, we find that the semantic similarity is weakly positively correlated with classification performance on Dog Story (r(3)=0.415, p=0.488) and strongly positively correlated with classification performance on ADReSS (r(3)=0.953, p=0.012). Further, the image description transformation has the lowest semantic similarity and lowest classification performance on both datasets. Overall, these seem to indicate that while text-to-text transformations are able to keep the relevant semantic indicators linked to AD, these are not maintained in text-to-image-to-text transformations. Whether this is due to the changes of modality or to differences in the quality of the text and image models remains to be investigated.

\subsection*{Further syntactic and lexical measures}
We now analyse additional metrics for quantifying the surface form differences between the texts produced by each transformation and the original data. We do this to understand the impact of the transformations on the data and whether the transformations also substantially change the distributional profile of the surface forms of the original texts.

\subsubsection*{Type-Token Ratio and Lexical Frequency}
There are a number of syntactic and lexical measures that have been shown in the literature to have different distributions in texts produced by patients with cognitive decline due to AD and study participants without AD. Due to the increased chance of repetition, it has been shown that the speech from AD patients has a lower type-token ratio (ttr) \cite{8217706, 8612831, mueller_connected_2016} and use of words with a higher lexical frequency (lf)  \cite{almor_why_1999}. These two measures have also been shown to be significant markers of AD when used as features for machine learning systems \cite{ben_ammar_speech_2018}, and have been used as evaluation metrics for text generated by language models as a response to picture description tasks \cite{li_gpt-d_2022}.

To evaluate to what extent these measures differ in the original texts from both groups (AD and Control) and if our transformations alter them, we calculate the average of these values in the original text and the transformations and measure if there are significant differences between them. Type-token ratio is calculated by taking the ratio of unique words to all words in the text, and lexical frequency is calculated using the average Zipf frequency taken from the wordfreq python library\cite{robyn_speer_2022_7199437}. We use Welch's t-test to compare the metrics between the control and AD groups, and report the p-value. We focus this analysis only on the ADReSS dataset, for comparative reasons with related work.

Comparing the original data for the two groups in the ADReSS dataset, neither of these measures is significantly different between the groups (Table \ref{tab:adress-lexical}, Original). Moreover, our transformations only introduce significant difference between the groups in 2 cases, lf in the short summaries and ttr in the image descriptions, and the absolute effect in both of these cases is rather small. Overall, neither metric provides a strong diagnostic signal, and our transformations do not affect this.

\begin{table}[!ht]
\centering
\caption{Lexical measures on the ADReSS dataset}
\label{tab:lexical-measures}
\begin{tabular}{l|lll|lll}
\toprule
Transformation & ttr$_C$ & ttr$_{AD}$ & ttr$_p$ & lf$_C$ & lf$_{AD}$ & lf$_p$ \\
\midrule
Original & 0.641 & 0.640 & 0.947 & 6.011 & 6.033 & 0.464 \\  \hdashline
Short Summaries & 0.819 & 0.825 & 0.589 & \textbf{5.759} & \textbf{5.814} & \textbf{0.042} \\
Medium Summaries & 0.813 & 0.827 & 0.222 & 5.763 & 5.791 & 0.343 \\
Long Summaries & 0.498 & 0.471 & 0.140 & 5.889 & 5.905 & 0.224 \\
Storyboard & 0.538 & 0.555 & 0.327 & 5.684 & 5.729 & 0.106 \\
Image Description & \textbf{0.722} & \textbf{0.735} & \textbf{0.024} & 5.708 & 5.684 & 0.163 \\
\bottomrule
\end{tabular}
\par\smallskip
\begin{flushleft}The mean measures for the control ($C$) and AD ($_{AD}$) on the ADReSS dataset and transformations. P-values ($p$) are given for the difference between the two groups. Any statistically significant differences ($p < 0.05$) are shown in bold.
\end{flushleft}
\label{tab:adress-lexical}
\end{table}

\subsubsection*{Part-of-Speech based metrics}

Other measures that have also been shown to be useful indicators of AD when analyzing language characteristics involve changes in distributions of certain parts-of-speech, specifically, with increased pronoun-to-noun (pnr) ratio \cite{bittner_changes_2022}, adverb ratio (RB), and participle ratio (VB) \cite{fraser_linguistic_2016} having been observed. The average value for each of these measures in the AD and control group for each of the translations, and the p-value (calculated using Welch's t-test) for the statistical significance of the differences are shown in Table \ref{tab:adress-pos}. Although there is a statistically significant difference in all of the measures in the original text (indicated by the p-values less than or equal to 0.001 for Original), there are very few statistically significant differences for our transformations. Additionally, in the cases of significance in the transformed texts, the absolute effect is much lower than that seen in the original texts. These results show that our transformations significantly reduce the diagnostic signal from part-of-speech based linguistic measures.


\begin{table}[!ht]
\begin{adjustwidth}{-2.25in}{0in}
\centering
\caption{Part-of-speech based measures on the ADReSS dataset}
\label{tab:pos-measures}
\begin{tabular}{l|lll|lll|lll}
\toprule
Transformation & pnr$_C$ & pnr$_{AD}$ & pnr$_p$ & RB$_C$ & RB$_{AD}$ & RB$_p$ & VB$_C$ & VB$_{AD}$ & VB$_p$ \\
\midrule
Original & \textbf{0.603} & \textbf{1.101} & \textbf{0.001} & \textbf{0.028} & \textbf{0.053} & \textbf{$<$0.001} & \textbf{0.095} & \textbf{0.077} & \textbf{$<$0.001} \\  \hdashline
Short Summaries & 0.130 & 0.163 & 0.252 & 0.015 & 0.022 & 0.056 & 0.117 & 0.111 & 0.454 \\
Medium Summaries & \textbf{0.163} & \textbf{0.232} & \textbf{0.044} & 0.016 & 0.023 & 0.090 & 0.112 & 0.108 & 0.574 \\
Long Summaries & 0.211 & 0.230 & 0.187 & 0.035 & 0.036 & 0.576 & 0.082 & 0.077 & 0.099 \\
Storyboard & 0.120 & 0.137 & 0.211 & \textbf{0.024} & \textbf{0.029} & \textbf{0.049} & 0.063 & 0.068 & 0.191 \\
Image Description & 0.201 & 0.208 & 0.688 & 0.039 & 0.040 & 0.877 & 0.095 & 0.097 & 0.485 \\
\bottomrule
\end{tabular}
\par\smallskip
\begin{flushleft} The mean measures for the control ($C$) and AD ($_{AD}$) on the ADReSS dataset and transformations. P-values ($p$) are given for the difference between the two groups. Any statistically significant differences ($p < 0.05$) are shown in bold.
\end{flushleft}
\label{tab:adress-pos}
\end{adjustwidth}
\end{table}

These analyses provide additional confirmation that our transformations significantly change the surface form  characteristics of the texts. This, paired with the high semantic similarity and the classification performance, getting close to or surpassing the original text for some of our transformations, shows that language models are still able to access the relevant latent indicators linked to AD even after the transformations applied to a patient's response.




\subsection*{Effect of translation}

When translating from a source to a target language, substantial changes in the surface form would be expected due to differences in conventionality between the two languages, with possible loss of information, which could affect performance for AD classification. On the other hand, the translation into a more resourced language, like English, allows access to potentially bigger models trained on larger quantities of data. Therefore, the impact of these changes needs to be determined. 

From our results for the Dog Story dataset, translating the original texts into English and using an English-only classification model increases the performance compared to processing the data in Portuguese (in Table \ref{tab:dog-back-translate} Original Portuguese vs Translated English). As we cannot directly compare the syntax across languages, it is hard to understand whether this improvement comes from the translation fundamentally changing the text, or it is a result of English language models performing better due to English being a more richly resourced language.

To test this, we explore translating the text back into Portuguese from the translated English data using the NLLB-200 model. We then train a classifier based on BERTimbau as we did with the original data. The results are in Table \ref{tab:dog-back-translate}.

\begin{table}[!ht]
\caption{\bf Back translation results on Dog Story dataset}
\centering
\begin{tabular}{lccc}
\toprule
Data Transformation & macro-F1 & Positive Accuracy & Negative Accuracy \\
\midrule
Original Portuguese & 0.616 & 0.283 & \textbf{0.923} \\
Translated English & \textbf{0.640} & \textbf{0.330} & 0.922 \\
Back Translated Portuguese & 0.634 & \textbf{0.330} & 0.914 \\
\bottomrule
\end{tabular}
\vspace{0.5em}
\\
\begin{flushleft}
The mean metrics across 10 runs. The best model for each metric is shown in bold.
\end{flushleft}
\label{tab:dog-back-translate}
\end{table}

Whilst the results are not as high as those achieved with the English data, they are significantly better than those achieved with the original Portuguese data. This suggests that there are performance gains from using the English language models. Moreover, the translation seems to fundamentally change the text, but in a way that improves access to the latent information that is relevant for AD.

A manual analysis of the data confirms that both the translations into English and back-translations to Portuguese are mostly good quality, with some small grammatical mistakes made by the model. The back translations appear to be more fluent, removing some of the markedness of the original texts, especially those of AD patients. Although we would expect this change to negatively affect the AD classification performance as the irregularities should be useful indicators, this may remove some of the confounding clues from the surface form and make detection of the latent semantic differences easier. Further work is needed to verify this, however.

\section*{Conclusion}

In this work, we examined how we can use large language models to modify the transcripts of patients with Alzheimer's Disease, such that the surface (lexico-syntactic) form is changed, whilst keeping the meaning (semantics) the same. Using a range of metrics, we show that our pipeline of text transformations successfully modifies the lexico-syntactic features of the texts, as measured with low BLEU and chrF scores, whilst maintaining high semantic similarity, as measured with cosine similarity from SentenceBERT. Across two datasets in English and Portuguese, we train BERT classifiers with both these modified texts and the originals, finding only small, both positive and negative, performance differences on Alzheimer's detection. These results contribute to our understanding of how Alzheimer's affects language production, suggesting that semantic degradation is the most fundamental effect of the progression of the disease, with lexico-syntactic degradation being a byproduct, supporting the conclusions of some prior work.

We also look at an approach which generates images from LLM-generated storyboards, and then captions them back into text, as a further way to create summaries. This appproach, however, leads to a reduction in performance, requiring further investigation.

Our work opens up the avenue for future research, in particular around the impairment of semantic language abilities and their role in automatic AD detection. Additionally, we envision that our pipeline could also facilitate de-identification of patients' transcripts, and the creation of augmented data that could expand the available training data for AD detection.

\section*{Acknowledgments}
\nolinenumbers

%
%
%
\bibliography{references,custom}

\newpage
\appendix
\renewcommand{\thesection}{S\arabic{section}}
\setcounter{section}{0}

\section{Example Pipeline Output}
\label{app:example-output}
\begin{table}[h!]
\begin{adjustwidth}{-2.25in}{0in}
\caption{
{\bf Example outputs from the pipeline}}
\begin{tabular}{c p{0.16\linewidth} p{0.72\linewidth}}
\toprule
Step & Transformation & Prompt \\
\midrule
0 & Original & O menino ia passando na rua e viu um cachorrinho. Gostou do cachorrinho, aí chamou o cachorrinho, e o cachorrinho veio do lado dele. Aí, o cachorrinho acompanhou ele, né? Aí, chegando em casa, ele escondeu o cachorrinho dentro do roupeiro pra sua mãe não ver. Mas sua mãe achou o cachorrinho dentro do roupeiro e brigou com o menino. O menino pediu, por favor, pra ela, e aí ela deixou. Tanto é que fez a casinha e tudo, ajudou ele a fazer a casinha pra colocar o cachorro.\\ \hdashline
1 & Translated & The boy was walking down the street and saw a puppy. He liked the puppy, so he called the puppy, and the puppy came to his side. Then the puppy followed him, didn't it? Then, coming home, he hid the puppy inside the wardrobe so his mother wouldn't see it. But his mother found the puppy inside the wardrobe and got into a fight with the boy. The boy asked her for it, and then she left. So he did the house and everything, helped him make the house to put the puppy.\\\hdashline
2 & Short Summary & A boy found a puppy he liked on the street, hid it from his mother who eventually discovered it, leading to a disagreement after which he helped make a place for the puppy at home. \\ \hdashline
2 & Medium Summary & The boy found a puppy on the street and brought it home, hiding it in a wardrobe to keep it from his mother. However, she discovered the puppy, leading to an argument. Eventually, the boy convinced her to let him keep the puppy, and they created a space for it at home. \\ \hdashline
2 & Long Summary & A boy spotted a puppy while walking and immediately took a liking to it. When he called the puppy, it obediently came to him and began to follow him. Wanting to keep the puppy a secret, the boy hid it inside a wardrobe upon reaching home. However, his mother soon discovered the puppy and confronted the boy, leading to a disagreement between them. Despite the initial conflict, the boy pleaded with his mother, who eventually relented. Consequently, the boy took on the responsibility of making a suitable home for the puppy, diligently preparing a space for it. \\ \hdashline
2 & Storyboard & 1. **The Encounter:** The boy walks down the street and spots a playful puppy. He feels an instant fondness for the little creature.

2. **The Bonding:** The boy calls out gently to the puppy, and it eagerly trots to his side, forming an immediate companionship.

3. **The Secret:** As they reach home, the boy decides to keep the puppy a secret and carefully hides it inside his wardrobe, hoping his mother won't notice.

4. **The Discovery:** The boy's mother unknowingly opens the wardrobe and is surprised to find the hidden puppy, leading to a confrontation with her son.

5. **The Conflict:** The boy pleads with his mother to keep the puppy, expressing his desire and attachment to it.

6. **The Resolution:** Touched by her son's plea, the mother decides to leave the room, allowing him to figure out how to care for the puppy.

7. **The New Beginning:** With determination, the boy takes on responsibilities, cleaning the house and creating a makeshift home to care for his new puppy, symbolizing his commitment and love. \\ \hdashline
3 & Image Caption & The image displays a red door with a white envelope tucked into the door jamb. This suggests a scenario where someone has left a message or letter for the person inside. The envelope, peeking out from the door, adds a sense of anticipation or communication, perhaps left for someone who isn't home. The area is otherwise devoid of people or additional actions, focusing solely on the simple, but potentially significant act of leaving a note. \\ 
\bottomrule
\end{tabular}
\begin{flushleft}
Example outputs from each stage of our pipeline for a randomly chosen example from the Dog Story dataset.
\end{flushleft} 
\label{tab:example-output}
\end{adjustwidth}
\end{table}

\clearpage
\section{Full similarity results}
\label{sec:conf-matrices}

\begin{figure}[!h]
\includegraphics[width=\textwidth]{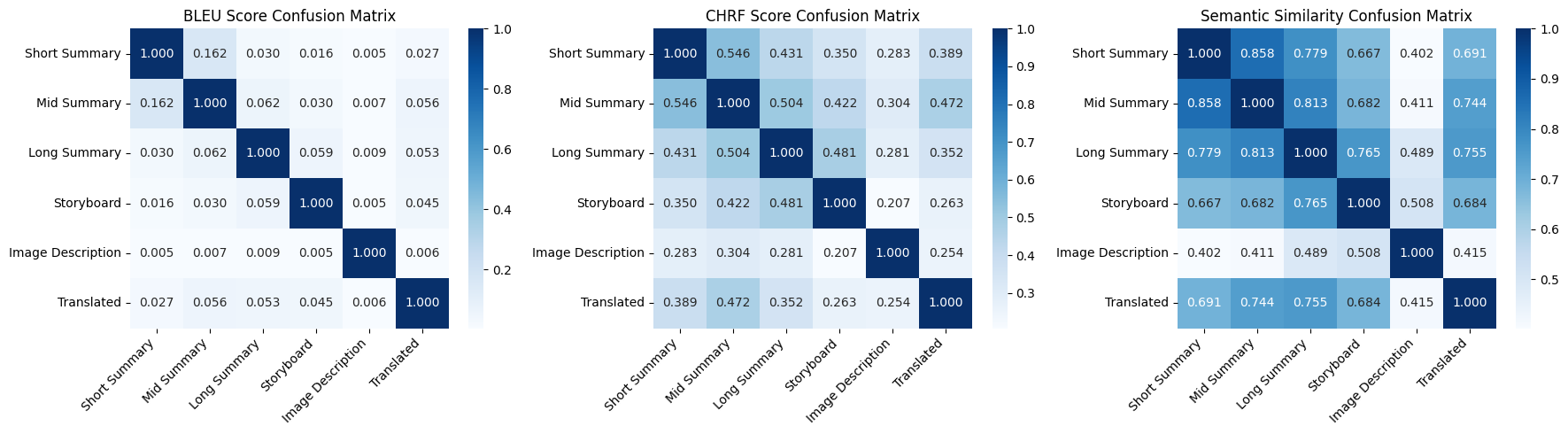}
\caption{{\bf Dog Story similarity metrics} Complete pairwise similarity scores for all the transformations of the Dog Story dataset, presented in confusion matrices.}
\label{fig:conf-dog}
\end{figure}

\begin{figure}[!h]
\includegraphics[width=\textwidth]{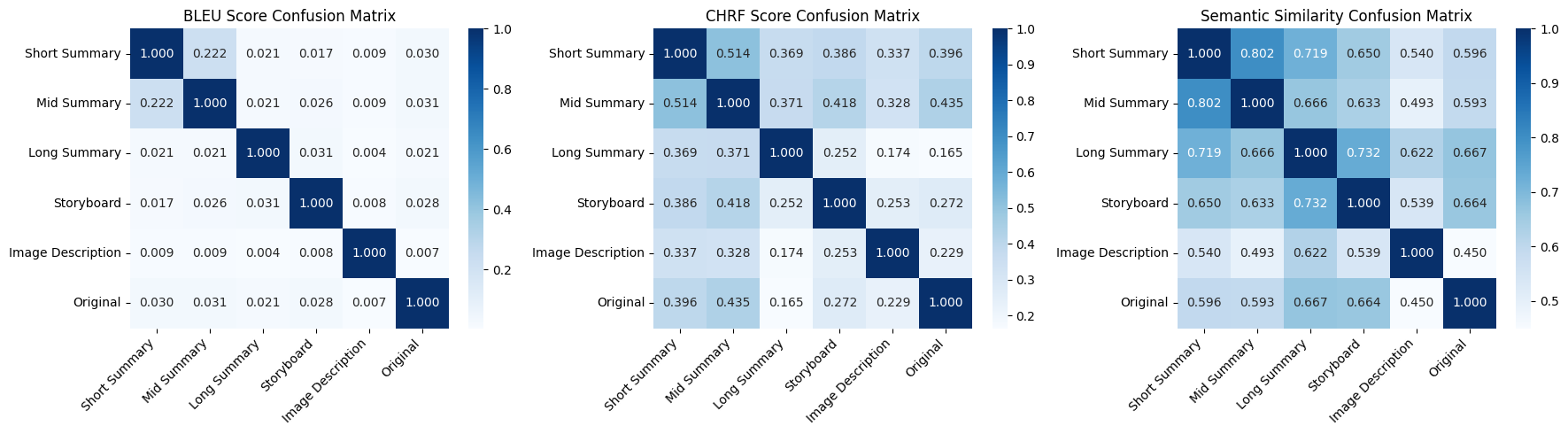}
\caption{{\bf ADReSS similarity metrics} Complete pairwise similarity scores for all the transformations of ADReSS, presented in confusion matrices.}
\label{fig:conf-adress}
\end{figure}

In addition to performing the similarity test between the original texts and the transformations, we also calculate the same metrics between all pairs of texts. These metrics are presented as confusion matrices in Fig \ref{fig:conf-dog} for the Dog Story dataset, and Fig \ref{fig:conf-adress} for ADReSS.

Overall, a similarly low level of syntactic similarity is seen between the transformed texts as was seen when comparing each transformation to the original data. This shows that our transformations produce diverse texts and strengthens our claims that the classification models are robust to changes in the surface form.

For semantic similarity, again, the pairwise similarities are high (with the exception of comparisons with the image caption) and in the range of scores between the transformed texts and the originals. This shows that the general meaning of the texts is being preserved across all the transformations, with some small variations in the information kept by the models. This may explain some of the variation in classifier performance observed.

\end{document}